\documentclass[a4paper,twoside]{article}

\usepackage{epsfig}
\usepackage{url}
\usepackage{subcaption}
\usepackage{calc}
\usepackage{amssymb}
\usepackage{amstext}
\usepackage{amsmath}
\usepackage{amsthm}
\usepackage{multicol}
\usepackage{pslatex}
\usepackage{apalike}
\usepackage{algorithm2e}
\usepackage[bottom]{footmisc}
\usepackage{SCITEPRESS}

\begin{document}

\title{LLMQuoter: Enhancing RAG Capabilities Through Efficient Quote Extraction From Large Contexts}

\author{\authorname{Yuri Façanha Bezerra\sup{1}, Li Weigang\sup{1} }
\affiliation{\sup{1}TransLab, University of Brasilia, Brasilia, Federal District, Brazil}
\email{yurifacanha1994@gmail.com, weigang@unb.br}
}


\abstract{We introduce LLMQuoter, a lightweight, distillation-based model designed to enhance Retrieval-Augmented Generation (RAG) by extracting the most relevant textual evidence for downstream reasoning tasks. Built on the LLaMA-3B architecture and fine-tuned with Low-Rank Adaptation (LoRA) on a 15,000-sample subset of HotpotQA, LLMQuoter adopts a ``quote-first-then-answer" strategy, efficiently identifying key quotes before passing curated snippets to reasoning models. This workflow reduces cognitive overhead and outperforms full-context approaches like Retrieval-Augmented Fine-Tuning (RAFT), achieving over 20-point accuracy gains across both small and large language models.
By leveraging knowledge distillation from a high-performing teacher model, LLMQuoter achieves competitive results in a resource-efficient fine-tuning setup. It democratizes advanced RAG capabilities, delivering significant performance improvements without requiring extensive model retraining. Our results highlight the potential of distilled quote-based reasoning to streamline complex workflows, offering a scalable and practical solution for researchers and practitioners alike.
}

\onecolumn \maketitle \normalsize \setcounter{footnote}{0} \vfill
\section{\uppercase{Introduction}}
Large Language Models (LLMs) have revolutionized natural language processing by enabling robust performance across diverse tasks such as open-domain question answering, summarization, and conversational AI. However, as these models grow in size and capability, their computational demands and inefficiencies in handling large contexts have become an active area of research, driving efforts to develop innovative strategies for improvement \cite{lin2024infinite,jin2024llm,an2024make}. 
This challenge is particularly pronounced in scenarios requiring complex reasoning and retrieval of specific information from extensive textual data.

Retrieval-Augmented Generation (RAG) has become a powerful paradigm for improving model performance by seamlessly integrating retrieval mechanisms with generative capabilities. By integrating external knowledge sources, RAG enables models to access and utilize relevant information dynamically, enhancing their domain-specific performance without necessitating extensive retraining. Despite its potential, RAG still faces limitations, especially in smaller models, which often struggle to reason effectively when confronted with large or noisy contexts \cite{mirzadeh2024gsm,hu2024minicpm}.

To address these limitations, we propose \textit{LLMQuoter}, a lightweight model designed to enhance RAG workflows by adopting a ``quote-first-then-answer'' strategy. Instead of reasoning over an entire context, LLMQuoter extracts relevant textual snippets that directly support downstream reasoning tasks. This approach reduces cognitive load on reasoning models, enabling both small and large models to achieve superior accuracy with lower computational overhead.

Our methodology builds on recent advances in knowledge distillation, where compact models are trained to emulate the capabilities of high-performing teacher models. By leveraging distillation techniques and fine-tuning a LLaMA-3B model with Low-Rank Adaptation (LoRA) \cite{hu2021lora}, we demonstrate that LLMQuoter can effectively identify and extract key quotes from large contexts. These extracted quotes are then passed to reasoning models, enabling a “quote-first-then-answer” strategy. This approach departs from traditional full-context techniques and frameworks like Retrieval-Augmented Fine-Tuning (RAFT) \cite{raft}, where the model quotes, thinks, and answers in a single step. By decoupling these stages, LLMQuoter simplifies the reasoning process and reduces cognitive overhead for downstream models, offering an alternative and efficient pathway for retrieval-augmented generation (RAG) pipelines.

This paper evaluates LLMQuoter using the DSPy framework \cite{khattab2023dspy} for semantic evaluation, leveraging a 15,000-sample subset of the HotpotQA dataset \cite{yang2018hotpotqa}. This dataset is commonly used for RAG problems and was also utilized in RAFT, the solution that served as our starting point. The results show that LLMQuoter is a standout solution, holding its own against RAG techniques like RAFT. It delivers impressive accuracy gains without compromising on computational efficiency, making it both effective and resource-friendly. Furthermore, the lightweight nature of LLMQuoter democratizes access to advanced RAG capabilities, providing a scalable solution for researchers and practitioners with limited resources.
The rest of this paper is organized as follows: Section 2 delves into the background, covering LLM reasoning, knowledge distillation, the RAFT approach, and evaluation methods for LLMs. Section 3 outlines the methodology behind the proposed solution. In Section 4, we detail the experimental setup and key insights from the process, while Section 5 discusses the results and highlights the advantages of quote-based reasoning. Finally, Section 6 wraps up the paper with conclusions and potential avenues for future research.

\section{\uppercase{Background}}
\subsection{LLM Reasoning}
Reasoning remains a core challenge for Large Language Models (LLMs), with both large and small models facing distinct limitations. Large models excel in generalization but struggle with intricate logical reasoning and multi-step problem-solving. Rather than genuinely reasoning, they often replicate patterns from their training data, which can lead to significant performance drops when faced with tasks requiring clause integration or minor input variations, as highlighted in the GSM-Symbolic study \cite{mirzadeh2024gsm}. Smaller models, while resource-efficient, suffer from capacity constraints, making them prone to losing context in reasoning-intensive tasks, as demonstrated in experiments with MiniCPM, which attempts to match the reasoning performance of larger models \cite{hu2024minicpm}.

To mitigate these challenges, split-step reasoning has emerged as a promising solution. By dividing reasoning tasks into distinct phases, such as problem decomposition and problem-solving, smaller models can focus on manageable subtasks, improving their generalization and inference efficiency \cite{wu2024divide}. Advanced techniques like Generative Context Distillation (GCD) and task-specific fine-tuning further enhance reasoning accuracy while preserving computational efficiency \cite{fu2024specializing}. This approach has also been successfully applied in specific domains such as sarcasm detection, where frameworks like chain-of-contradictions outperform holistic reasoning approaches, particularly in smaller models \cite{yao2024sarcasm}.

Self-correction mechanisms provide an additional boost to reasoning capabilities. Training pipelines that incorporate self-correction, where models generate critiques of their incorrect answers, have proven effective, particularly when pairing small models with strong verifiers \cite{zhang2024selfcorrection}. Domain-specific approaches, such as multi-modal assistants integrating textual and visual reasoning, further demonstrate that smaller models can achieve sophisticated reasoning abilities when tailored strategies are employed \cite{zhu2024llava}.

These studies underscore the importance of split-step reasoning and specialized training frameworks to address the limitations of both large and small LLMs. By leveraging strategies that combine task decomposition, fine-tuning, and self-correction, researchers can design models that effectively balance scalability and reasoning performance across diverse applications.

\subsection{Knowledge Distillation in LLMs}

Knowledge distillation has become a vital technique for reducing the computational demands of large language models (LLMs) while preserving their advanced capabilities. The process transfers knowledge from a high-capacity teacher model to a more efficient student model, enabling smaller models to perform complex tasks such as reasoning, recommendation, and maintaining factual accuracy, all with significantly lower resource consumption. Techniques like Generative Context Distillation (GCD), for instance, streamline inference by internalizing prompts rather than relying on explicit ones, enhancing efficiency \cite{fu2024gcd}. Similarly, rationale-based approaches help compact models achieve state-of-the-art performance in recommendation tasks by improving model profiling \cite{hu2024rdrec}.

The applications of LLM distillation are diverse, ranging from mitigating hallucinations to enhancing recommendation systems and enabling cost-effective deployment. Techniques such as temperature scaling and intermediate layer matching have been shown to reduce hallucination rates while improving accuracy \cite{gogate2024techrxiv}. In the domain of recommendation systems, rationale-based and importance-aware ranking distillation techniques allow smaller models to effectively learn from user-item interactions, striking a balance between computational efficiency and performance \cite{cui2024dllm2rec}. Task-specific strategies, such as dividing reasoning tasks into problem decomposition and solution phases, further improve the generalizability and inference efficiency of compact models \cite{wu2024divide}.

Despite its advantages, knowledge distillation faces challenges, including capacity gaps between teacher and student models and semantic divergence in embedding spaces. Solutions like collaborative embedding distillation and ranking distillation have been developed to bridge these gaps, allowing smaller models to align more closely with their larger counterparts in semantic reasoning \cite{cui2024dllm2rec}. As researchers continue to optimize these methods, they push the boundaries of LLM distillation, making smaller, efficient models increasingly viable for a broad spectrum of AI applications.

\subsection{RAFT (RAG + FT)}

Retrieval-Augmented Generation (RAG) has emerged as a widely adopted technique for domain-specific applications, offering an efficient way to leverage pre-trained LLMs without requiring extensive retraining. RAG is particularly advantageous as it allows models to retrieve relevant information from external knowledge bases or documents, making it an effective approach in fields such as healthcare, legal analysis, and technical documentation, where large, specialized datasets are required  \cite{rag_for_healthcare,legal_bot}. Additionally, RAG is increasingly being utilized as an external memory mechanism for LLMs \cite{memory_Rag,memoryhipporag}.

However, not all models—especially smaller ones—are capable of effectively handling large contexts and reasoning simultaneously \cite{zhang2024selfcorrection,i_dont_know_ai}. Even when provided with the appropriate context, these models often fail to generate coherent or accurate answers, revealing a gap in their ability to integrate retrieval with reasoning. To address these limitations, techniques like RAFT, Retrieval-Augmented Fine-Tuning, have emerged, aiming to fine-tune LLMs specifically for RAG in domain scenarios \cite{raft,dioliveira2024slim}. 

\begin{figure}[!h]
  \centering
    \includegraphics[width=0.45\textwidth]{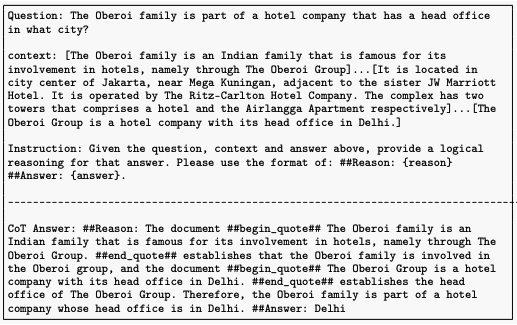}
  \caption{RAFT inference example\cite{raft}}
  \label{fig:raft_example}
 \end{figure}

RAFT focuses on training models to ``think while quoting," combining reasoning with the ability to extract and reference relevant portions of the retrieved context. This dual focus enables the model to dynamically identify key parts of the input text, synthesize their meaning, and produce a well-reasoned final answer. By teaching models to reason and quote in tandem, RAFT enhances their ability to operate effectively in RAG settings, bridging the gap between retrieval and generation to deliver more accurate, context-aware responses. In summary, RAFT trains a model to perform chain-of-thought (CoT) reasoning, transforming a task structured as context + question into a comprehensive output of reasoning, relevant quotes, and the final answer (Figure \ref{fig:raft_example}).

\subsection{Semantic Evaluation}

Semantic evaluation of Large Language Models (LLMs) has emerged as a critical area of research as these models continue to excel in tasks such as text summarization, question answering, and open-domain generation. With their growing sophistication, there has been a corresponding rise in prompt-based systems designed work with LLM as the product brain \cite{smantic_ex1,smantic_ex2,smantic_ex4}. However, evaluating the effectiveness of LLM-generated outputs objectively remains a significant challenge \cite{hu2024llmevaluation}. Traditional metrics often fall short in capturing the nuanced semantics and creative aspects of these models’ outputs, necessitating the development of more refined evaluation frameworks \cite{khattab2023dspy}.

Traditional metrics such as BLEU and ROUGE are limited by their reliance on surface-level token overlaps, failing to capture the semantic depth and creativity of LLM-generated outputs\cite{van2024field}. This limitation has driven the development of more robust and adaptable evaluation frameworks, including LLMs themselves as evaluators. Frameworks like GPTScore \cite{gptscore} and AlpacaEval \cite{alpacaeval} exemplify this shift: GPTScore offers granular scoring based on conditional probabilities but requires access to token-level data; and AlpacaEval simplifies comparative evaluation with a win-rate metric, while also exposing challenges like prompt sensitivity.

DSpy and EvalGen represent significant advancements in automating and structuring LLM evaluations. EvalGen combines automation with human feedback to iteratively refine evaluation criteria, addressing the evolving nature of semantic assessment and introducing the concept of \textit{criteria drift}, where grading outputs helps refine standards. DSpy adopts a “programming, not prompting” philosophy, enabling the programmatic implementation of metrics and evaluations using another LLM as a semantic judge. This approach abstracts the complexity of prompt engineering, allowing for reusable and scalable evaluations across tasks. Together, these frameworks address the limitations of traditional metrics and lay the groundwork for a rigorous, nuanced, and adaptable evaluation landscape for LLMs.

\section{\uppercase{Methodology}}

With the goal of developing an efficient language model for extracting relevant quotes from contexts to properly answer questions about it, this section details the methodology employed in training and evaluating the distilled LLM. The process involves leveraging a high-performing LLM for dataset creation, fine-tuning a smaller LLM, and validating the approach with task-specific metrics. 

We begin with a formalization of the distillation problem in Section 3.1, followed by an overview of the fine-tuning process in Section 3.2. Finally, the evaluation framework and metrics used to validate the model's performance are described, along with a simple approach to demonstrate the benefits of extracting relevant quotes instead of using the large content itself.

\subsection{Problem Formalization}
Let us consider a dataset of text samples, denoted by \( D = \{(C, Q, A)\} \), where:
\begin{itemize}
    \item \( C \): a large text context.
    \item \( Q \): a specific question.
    \item \( A \): the expected answer.
\end{itemize}
The task is to train a model capable of extracting relevant quotes from \( C \) that support \( A \) in response to \( Q \). 

To achieve this, we employ a distillation process in which a large LLM generates high-quality training data, and a smaller LLM is fine-tuned on this dataset to efficiently replicate the behavior of the larger model.

\subsection{LLM Distillation}
The dataset creation process can be formalized as follows: Given a high-performance language model \( f_{\text{high}} \), such as ChatGPT or Gemini, the task is to extract quotes \( \mathcal{R} \) from a context \( C \) that directly support an answer \( A \) in response to a question \( Q \). Formally, this process can be represented as: 
\[
f_{\text{high}}: (Q, A, C) \rightarrow \mathcal{R}
\]
For each data point \( (Q, A, C) \), the high-performance model \( f_{\text{high}} \) generates the set of quotes \( \mathcal{R} \), which serve as the ground truth: 
\[
\mathcal{D}_{\text{gold}} = \{(Q, A, C, \mathcal{R}) \mid \mathcal{R} = f_{\text{high}}(Q, A, C)\}
\]
The result is a high-quality dataset \( \mathcal{D}_{\text{gold}} \), consisting of tuples \( (Q, A, C, \mathcal{R}) \), where \( \mathcal{R} \) represents the relevant quotes extracted by \( f_{\text{high}} \). This dataset is then used to train and evaluate the smaller distilled model \( f_{\text{small}} \).

\subsection{Fine-Tuning LLM with LoRA}
The smaller model \( f_{\text{small}} \) is fine-tuned on the \( \mathcal{D}_{\text{gold}} \) dataset using Low-Rank Adaptation (LoRA) for task-specific learning in the extraction of relevant quotes. The fine-tuning process is defined as:
\[
f_{\text{small}}: (Q, C) \rightarrow \mathcal{R}
\]
where \( Q \) represents the question, \( C \) is the textual context, and \( \mathcal{R} \) is the set of relevant quotes generated by the fine-tuned model. The training process is described in the following steps:
\begin{enumerate}
    \item \textbf{Input:} Data from the \( \mathcal{D}_{\text{gold}} \) dataset in the form of tuples \( (Q, C) \), where \( Q \) is the question, \( C \) is the textual context.
  \item \textbf{Output:} The fine-tuned model \( f_{\text{small}} \) is optimized to predict \( \mathcal{R} \), replicating the behavior of the larger model \( f_{\text{high}} \), \textbf{but without knowing the answer}.
   
\end{enumerate}

\subsection{Evaluation Framework and Metrics}
The model’s performance is evaluated using the DSpy framework, which computes task-specific metrics tailored to LLM outputs. Precision and recall are redefined for the quote extraction task using an LLM Judge to assess semantic relevance between model predictions and ground truth.

Precision measures the proportion of predicted quotes (\( R_{\text{model}} \)) that align semantically with the golden answers (\( R_{\text{gold}} \)), defined as:
\[
P = \frac{\sum_{r \in R_{\text{model}}} \text{Judge}(r, R_{\text{gold}})}{|R_{\text{model}}|}
\]
where \( R_{\text{model}} \) is the set of quotes predicted by the model, \( R_{\text{gold}} \) is the set of golden answers, and \( \text{Judge}(r, R_{\text{gold}}) \) is a scoring function returning values from 0 (no match) to 1 (perfect match).

Recall quantifies the proportion of golden answers (\( R_{\text{gold}} \)) captured by the model’s predictions (\( R_{\text{model}} \)), defined as:
\[
R = \frac{\sum_{r \in R_{\text{gold}}} \text{Judge}(r, R_{\text{model}})}{|R_{\text{gold}}|}
\]

F1-score balances precision and recall and is defined as:
\[
F_1 = 2 \cdot \frac{P \cdot R}{P + R}
\]

\paragraph{DSpy-Assisted Validation with LLM Judge:}  
The DSpy framework incorporates large language models (LLMs) as automated evaluators, enabling robust and interpretable metric calculations. This flexibility allows DSpy to integrate a wide range of LLMs, referred to here as the LLM Judge. This variation of precision and recall, tailored for LLM-generated outputs and supported by the LLM Judge’s semantic judgment, ensures a nuanced evaluation of the quote extraction model. The integration of DSpy and the Judge provides a systematic, interpretable, and robust framework for assessing and iteratively improving model performance.

\subsection{Proving the Benefit of Using Quotes}

Let \( f_{\text{base}} \) represent base models without any fine-tuning to establish a baseline for comparison. Two experimental setups are defined to demonstrate the advantage of using relevant quotes \( \mathcal{R} \) instead of the full context \( C \):

\begin{enumerate}
    \item Providing only the gold quotes \( \mathcal{R} \) from \( \mathcal{D}_{\text{gold}} \) to the base models \( f_{\text{base}} \) to answer the questions:
    \[
    f_{\text{base}}: (Q, \mathcal{R}_{\text{gold}}) \rightarrow A_{\text{base}}
    \]
    \item Providing the full context \( C \) instead of the quotes \( \mathcal{R} \) to the same base models \( f_{\text{base}} \) to answer the questions:
    \[
    f_{\text{base}}: (Q, C) \rightarrow A_{\text{base}}
    \]
\end{enumerate}

For both setups, \( Q \) represents the question, \( \mathcal{R}_{\text{gold}} \) is the set of gold quotes extracted from the \( \mathcal{D}_{\text{gold}} \) dataset, \( C \) is the entire context, and \( A_{\text{base}} \) is the base models answers. 

The accuracy of the answers produced by \( f_{\text{base}} \) is measured using Semantic Accuracy (\( S_{\text{acc}} \)), which evaluates the alignment between the model-generated answers \( A_{\text{base}} \) and the expected answers \( A_{\text{gold}} \). Semantic Accuracy is defined as:
\[
S_{\text{acc}} = \frac{\sum_{a \in A_{\text{base}}} \text{Judge}(a, A_{\text{gold}})}{|A_{\text{gold}}|}
\]
where \( \text{Judge}(a, A_{\text{gold}}) \) is a semantic similarity function scoring the alignment between a model-generated answer \( a \) and the ground truth \( A_{\text{gold}} \), with scores ranging from 0 (no match) to 1 (perfect match).

\section{\uppercase{Experiments}}

\begin{figure*}[!h]
  \centering
   {\epsfig{file=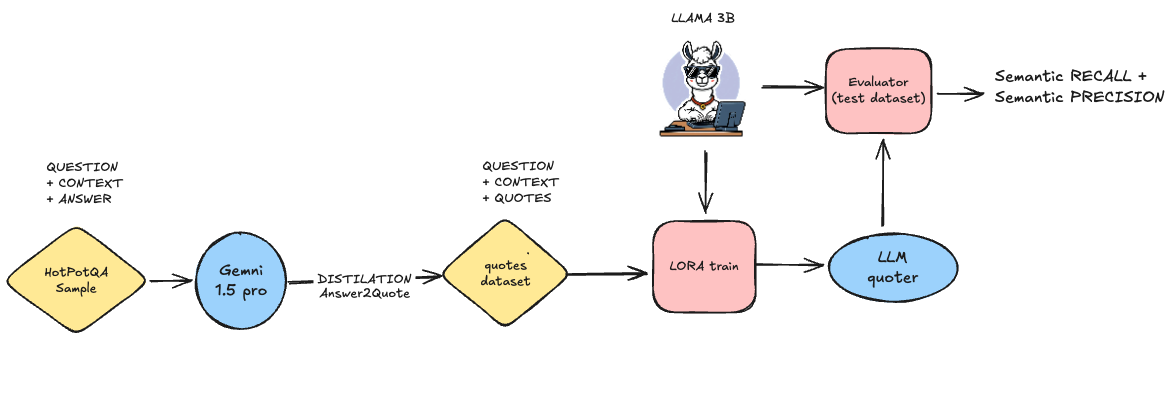, width = 1\textwidth}}
  \caption{The LLMQuoter diagram}
  \label{fig:fig1_quoter}
 \end{figure*}

\begin{figure}[!h]
  \centering
   {\epsfig{file=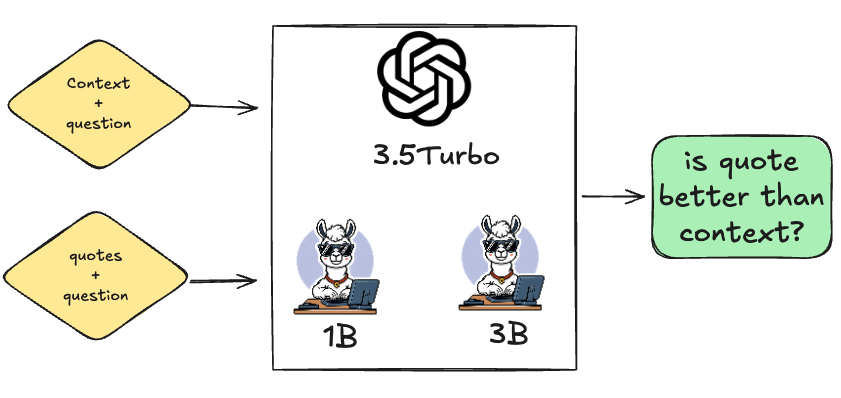, width = 7.5cm}}
  \caption{Context X Quotes process }
  \label{fig:comparison}
 \end{figure}

This section describes the experimental setup used to analyze the performance of the proposed methodology. It begins with details of the datasets used for training and evaluation, followed by an explanation of the training configurations, including hyper-parameters and computational resources. An overview of the entire process, from data distillation to evaluation, is illustrated in Figure~\ref{fig:fig1_quoter}. Finally, the experiments designed to validate the effectiveness of using relevant quotes instead of full context are presented (Figure \ref{fig:comparison} illustrates the process). The code utilized in this work is available on GitHub\footnote{https://github.com/yurifacanha/LLMQuoter}. Concrete examples of the experimental results can be found in the appendix for further clarification.

\subsection{Datasets}

Our method was evaluated on the HotpotQA dataset (Yang et al., 2018), an open-domain question-answering benchmark derived from Wikipedia, with a focus on common knowledge topics such as movies, sports, and general trivia. The dataset consists of three columns: \textbf{question}, \textbf{context}, and \textbf{answer}, where each sample pairs a question with a large textual context and its corresponding answer.

Due to resource constraints, a random subset of 15,000 samples was selected from the original dataset to serve as the basis for applying the distillation process. From this subset, 600 samples were set aside for evaluation purposes, forming the test set. This test set was used to measure the model's performance during the evaluation phase and to validate the benefit of using extracted quotes as opposed to the entire context for answering questions. The remaining 14,400 samples were utilized for training and validation during the distillation and fine-tuning steps.

\begin{table}[h]
\caption{Summary of dataset characteristics used in the experiments.}\label{tab:hotpotqa_dataset}
\centering
\begin{tabular}{|c|c|}
  \hline
  \textbf{Attribute} & \textbf{Value} \\
  \hline
  Dataset Name & HotpotQA \\
  \hline
  Total Samples Used & 15,000 \\
  \hline
  Test Set Size & 600  \\
  \hline
  Training Size & 14,400 \\
  \hline
  Source & Wikipedia \\
  \hline
  Topics & Common knowledge\\
 \hline
\end{tabular}
\end{table}

\subsection{Data Distillation}

The distillation process was performed using \textbf{Gemini Pro 1.5} as the high-performance model (\( f_{\text{high}} \)) and LangChain as the framework for managing the pipeline. The process involved generating relevant quotes for each sample in both the training and test datasets by leveraging the capabilities of Gemini Pro 1.5. 

Gemini Pro 1.5, as one of the most powerful models available today, was tasked with extracting quotes directly supporting the answer to each question. Given the model's advanced performance and ability to generate high-quality answers, it is reasonable to assume that the resulting dataset represents an excellent "gold" standard for the task of quote extraction. 

After this step, the dataset was finalized, augmented with a new column containing the extracted quotes (\( \mathcal{R} \)). This enriched dataset, now comprising question (\( Q \)), context (\( C \)), and quotes (\( \mathcal{R} \)), served as the foundation for training and evaluating the smaller \( f_{\text{small}} \) model.

\subsection{Fine-Tuning Process}

The fine-tuning process was applied to the smaller LLM, LLAMA 3.2 3B, using the Low-Rank Adaptation (LoRA) technique to optimize the model for the quote extraction task. LLAMA 3.2 3B was chosen as the base model due to its balance between computational efficiency and task-specific adaptability. The fine-tuning process was completed over a single epoch, ensuring efficient adaptation without overfitting. 

The fine-tuning process was conducted on a NVIDIA A100-SXM4-40GB GPU, with a maximum memory capacity of 39.564 GB. The specific resource utilization and training parameters are summarized below:

\begin{table}[h]
\caption{Summary of Fine-Tuning Configuration and Resource Usage}\label{tab:fine_tuning}
\centering
\begin{tabular}{|c|c|}
\hline
\textbf{Configuration/Metric}      & \textbf{Value}                            \\ \hline
Memory Usage                       & 3.56GB(peak)          \\ \hline
Training Memory                    & 1.06GB(peak)    \\ \hline
Batch Configuration                & Batch size: 2                  \\ \hline
Gradient accumulation steps                                  &  4            \\ \hline
Total effective batch size                                   &  8             \\ \hline
Training Steps                     & 60                 \\ \hline
Trainable Parameters               & 24M aprox    \\ \hline
Training Time                      & 5 minutes                                 \\ \hline
\end{tabular}
\end{table}

This setup highlights the efficiency of the LoRA approach in adapting a compact model like LLAMA 3.2 3B for specific tasks with minimal resource usage and rapid training over just one epoch (see Table ~\ref{tab:fine_tuning}).

\subsection{Evaluation and Proving the Benefits}

The evaluation of the extracted quotes was performed using the DSpy framework in conjunction with OpenAI GPT-4.0. GPT-4.0 was selected as it operates outside the scope of the training data and methods, is recognized as one of the top reasoning models, and remains unbiased regarding the problem context. By leveraging these tools, the metrics defined in the methodology section were concretely implemented and materialized for evaluating the system's performance in a structured and measurable way.

The following signature, referencing the DSpy documentation \cite{khattab2023dspy}, directly implements the precision and recall metrics defined in the methodology section, providing a clear framework for evaluating how well the extracted quotes align with the ground truth:

\begin{small}
\begin{verbatim}
class QuotesPrecisionRecall(dspy.Signature):
    """
    There are quotes from the 
    ground truth and quotes from 
    the system response.
    You must calculate the recall 
    and precision of 
    the system response.
    """
    ground_truth: str = dspy.InputField()
    system_response: str = dspy.InputField()
    recall: float = dspy.OutputField(desc="""
    fraction (out of 1.0) how much quotes 
    from the ground truth are present 
    in the system response"""")
    precision: float = dspy.OutputField(desc=
    """fraction (out of 1.0) how much 
    quotes from the system response 
    are present in the ground truth""")
\end{verbatim}
\end{small}

To validate the benefit of using quotes instead of the full context, comparisons were performed across several base models (\( f_{\text{base}} \)), including \textbf{LLAMA 3.2:1B, LLAMA 3.2:3B, GPT-3.5 Turbo}. These models were evaluated in two configurations: using extracted quotes \( \mathcal{R} \) and using the full context \( C \). The accuracy of the answers produced by these models was assessed to determine the effectiveness of the quote extraction approach. GPT-4.0 was chosen as the external LLM Judge again to compute Semantic Accuracy (\( S_{\text{acc}} \)).

\section{\uppercase{Results and Discussion}}

This section presents the experimental results obtained by evaluating the quote extraction model (quoter) and validating the benefit of using quotes over full context in open-domain question-answering tasks. The results demonstrate the effectiveness of the proposed method in improving the performance of both small and large language models in RAG (retrieval-augmented generation) scenarios.

\subsection{Evaluation of the Quoter Model}

The performance of the quoter model was evaluated using the metrics described in Section 4.3. The recall, precision, and F1-score were measured both before and after fine-tuning the smaller LLM using the LoRA approach. The results are summarized in Table \ref{tab:quoter_performance}.

\begin{table}[h]
\caption{Performance of the Quoter Model Before and After Fine-Tuning}\label{tab:quoter_performance}\centering
\begin{tabular}{|c|c|c|c|}
\hline
Metric  & Before & After  \\ \hline
Recall               & 48.3\%                       & 68.0\%(\textbf{+19.7\%})               \\ \hline
Precision            & 43.6\%                       & 71.0\%(\textbf{+27.4\%})               \\ \hline
F1-Score             & 41.3\%                      & 69.1\%(\textbf{+27.8\%})               \\ \hline
\end{tabular}
\end{table}

The results show significant improvements in all three metrics after fine-tuning the quoter model. The F1-score increased from 41.3\% to 69.1\%, demonstrating the quoter's ability to accurately identify relevant quotes with low computational resources and a compact model.

\subsection{Benefits of Using Quotes Over Full Context}

To validate the benefit of using quotes instead of full context, a comparison was performed using original models without any training. Both the gold quotes and the full context were provided as inputs to different models: LLAMA 1B, LLAMA 3B, and GPT-3.5 Turbo. The accuracy of the answers generated by each model in these two configurations is summarized in Table \ref{tab:quotes_vs_context}.

\begin{table}[h]
\caption{Comparison of Accuracy Between Using Full Context and Quotes}\label{tab:quotes_vs_context}
\centering
\begin{tabular}{|c|c|c|}
\hline
Model     &Context & Quotes \\ \hline
LLAMA 1B              & 24.4\%                     & 62.2\% (\textbf{+37.8\%})                    \\ \hline
LLAMA 3B              & 57.7\%                     & 83.0\% (\textbf{+25.3\%})                   \\ \hline
GPT-3.5 Turbo         & 75.8\%                     & 88.5\% (\textbf{+12.7\%})                  \\ \hline
\end{tabular}
\end{table}

The results highlight a clear improvement in accuracy when using gold quotes compared to full context. For instance, LLAMA 1B achieved an accuracy of 62.2\% with quotes versus 24.4\% with full context, and GPT-3.5 Turbo achieved 88.5\% with quotes versus 75.8\% with full context. These findings indicate that providing a good quoter model can significantly enhance the performance of both small and large language models in RAG scenarios.

\subsection{Discussion}

The results validate the hypothesis that using extracted quotes instead of full context significantly improves model performance in open-domain question-answering tasks. This finding aligns with the original RAFT approach, which involves reasoning and answering directly over the full context. However, our experiments demonstrate that separating the tasks—first extracting quotes with a simple quoter and then reasoning over the concise data—can lead to comparable or better outcomes with lower computational overhead.

\begin{table}[h]
\caption{Comparison of RAFT and Full Context Results on LLaMA2-7B over HotPotQA dataset}\label{tab:raft_results}
\centering
\begin{tabular}{|l|c|}
\hline
\textbf{Method}             & \textbf{Accuracy} \\ \hline
LLaMA2-7B + Full Context    & 26.43\%                 \\ \hline
RAFT (LLaMA2-7B)            & 35.28\%                \\ \hline
\end{tabular}
\end{table}

To provide context, RAFT was tested with LLaMA2-7B over the full dataset, achieving an accuracy of 35.28\% when reasoning over both context and question simultaneously. Using the same model (LLaMA2-7B) with only the full context reduced performance to 26.43\% (see Table \ref{tab:raft_results}). While our experiments used a random sample of 15,000 rows from the HotpotQA dataset due to resource constraints, the results are promising. For instance, even with a lightweight 3B quoter model fine-tuned with minimal resources on Colab, the quote-based approach significantly boosted accuracy for various downstream models.

The comparison highlights that the quoter technique is a promising alternative. By offloading the task of quote extraction to a small and efficient model, we can streamline the reasoning process for larger models, avoiding the pitfalls of over-reasoning. The "divide and conquer" strategy allows each model to focus on its strength: smaller models specialize in targeted preprocessing, while larger models excel in reasoning over concise, relevant data. 

While our study only utilized a subset of the HotpotQA dataset, the results suggest that the quoter technique offers a scalable and efficient solution for enhancing retrieval-augmented generation (RAG) pipelines. Notably, the models used with the extracted quotes were not fine-tuned to reason better, yet still achieved significant improvements in accuracy. This highlights the power of the quoter approach in simplifying the reasoning task by reducing the cognitive load on base models, allowing even non-optimized models to perform effectively.

This approach could serve as a viable alternative to RAFT in scenarios with limited resources, demonstrating that a well-trained quoter can democratize access to high-performing NLP solutions. By offloading the preprocessing task of identifying relevant information, the quoter enables base models to focus their reasoning capabilities on concise, relevant data rather than processing large and noisy contexts.

\section{\uppercase{Conclusions and Future Work}}

This study demonstrates the effectiveness of data distillation and lightweight training for enhancing Retrieval-Augmented Generation (RAG) systems. By leveraging a high-performing teacher model to distill relevant quotes and fine-tuning a compact model, we achieved significant improvements in model performance. The fine-tuning process required minimal resources, with just 5 minutes of training on an NVIDIA A100 GPU, yet delivered robust results.

The experiments validate that an efficient quoter model can substantially enhance RAG performance by reducing the cognitive load on the reasoning process. By focusing the model’s efforts on the answer rather than processing and reasoning over large contexts, we eliminate the need for extensive training while improving accuracy. This approach aligns with the principle of “divide and conquer,” where the reasoning task is simplified and made more manageable for even small models. Ultimately, our results demonstrate that high-quality quote extraction can democratize access to high-performing RAG capabilities across a range of computational constraints.

While this work has established a strong foundation for quote-based RAG, several avenues for future research remain open:
\begin{itemize}
    \item \textbf{Expanded Datasets:} Testing the methodology on a wider range of datasets, including those with different domains and levels of complexity, and utilizing larger samples from each dataset will provide a more comprehensive evaluation of the approach.
    \item \textbf{Reinforcement Learning:} Incorporating reinforcement learning techniques, such as Proximal Policy Optimization (PPO) or Direct Preference Optimization (DPO), could further refine the quote extraction and reasoning steps, enhancing the overall system performance.
    \item \textbf{Larger Models:} Training larger models, such as an 8B parameter LLAMA, could offer insights into the scalability of the proposed methodology and the impact of model size on RAG effectiveness.
    \item \textbf{Prompt Engineering:} Developing advanced prompt engineering techniques could optimize the quote extraction and reasoning processes, improving both accuracy and efficiency.
    \item \textbf{Applications Beyond RAG:} The methodology can be extended to other use cases, such as memory-augmented RAG systems, where the quoter serves as a lightweight mechanism for managing and retrieving relevant information from large, external knowledge bases.
\end{itemize}

By exploring these directions, we aim to further refine the quote-based RAG pipeline and expand its applicability to broader NLP tasks, offering scalable and resource-efficient solutions for both research and real-world scenarios.

\bibliographystyle{apalike}
{\small
\bibliography{main}}

\begin{thebibliography}{}

\bibitem[Alonso et~al., 2024]{memory_Rag}
Alonso, N., Figliolia, T., Ndirango, A., and Millidge, B. (2024).
\newblock Toward conversational agents with context and time sensitive long-term memory.
\newblock {\em arXiv preprint arXiv:2406.00057}.

\bibitem[An et~al., 2024]{an2024make}
An, S., Ma, Z., Lin, Z., Zheng, N., and Lou, J.-G. (2024).
\newblock Make your llm fully utilize the context.
\newblock {\em arXiv preprint arXiv:2404.16811}.

\bibitem[Chen et~al., 2024]{i_dont_know_ai}
Chen, X., Wang, L., Wu, W., Tang, Q., and Liu, Y. (2024).
\newblock Honest ai: Fine-tuning" small" language models to say" i don't know", and reducing hallucination in rag.
\newblock {\em arXiv preprint arXiv:2410.09699}.

\bibitem[Cui et~al., 2024]{cui2024dllm2rec}
Cui, Y., Liu, F., Wang, P., et~al. (2024).
\newblock Dllm2rec: Distillation from llms to sequential recommenders.
\newblock {\em arXiv preprint arXiv:2405.00338}.

\bibitem[Di~Oliveira et~al., 2024]{dioliveira2024slim}
Di~Oliveira, V., Bezerra, Y.~F., Weigang, L., Brom, P.~C., and Celestino, V.~R. (2024).
\newblock Slim-raft: A novel fine-tuning approach to improve cross-linguistic performance for mercosur common nomenclature.
\newblock In {\em WEBIST}.

\bibitem[Dubois et~al., 2024]{alpacaeval}
Dubois, Y., Galambosi, B., Liang, P., and Hashimoto, T.~B. (2024).
\newblock Length-controlled alpacaeval: A simple way to debias automatic evaluators.
\newblock {\em arXiv preprint arXiv:2404.04475}.

\bibitem[Fu et~al., 2023]{gptscore}
Fu, J., Ng, S.-K., Jiang, Z., and Liu, P. (2023).
\newblock Gptscore: Evaluate as you desire.
\newblock {\em arXiv preprint arXiv:2302.04166}.

\bibitem[Fu et~al., 2024a]{fu2024gcd}
Fu, J., Ng, S.-K., Jiang, Z., and Liu, P. (2024a).
\newblock Generative context distillation.
\newblock {\em arXiv preprint arXiv:2411.15927}.

\bibitem[Fu et~al., 2024b]{fu2024specializing}
Fu, J., Ng, S.-K., Jiang, Z., and Liu, P. (2024b).
\newblock Specializing smaller language models towards multi-step reasoning.
\newblock {\em arXiv preprint arXiv:2301.12726}.

\bibitem[Garcia et~al., 2024]{smantic_ex2}
Garcia, B., Westerfield, L., Yelemali, P., and Gogate, N. (2024).
\newblock Improving automated deep phenotyping through large language models using retrieval augmented generation.
\newblock {\em medRxiv}.

\bibitem[Gogate et~al., 2024]{gogate2024techrxiv}
Gogate, N. et~al. (2024).
\newblock Reducing llm hallucination using knowledge distillation: A case study with mistral large and mmlu benchmark.
\newblock {\em TechRxiv}.

\bibitem[Hu et~al., 2021]{hu2021lora}
Hu, E.~J., Shen, Y., Wallis, P., Allen-Zhu, Z., Li, Y., Wang, S., Wang, L., and Chen, W. (2021).
\newblock Lora: Low-rank adaptation of large language models.
\newblock {\em arXiv preprint arXiv:2106.09685}.

\bibitem[Hu et~al., 2024a]{hu2024minicpm}
Hu, S., Tu, Y., Han, X., et~al. (2024a).
\newblock Minicpm: Unveiling the potential of small language models with scalable training strategies.
\newblock {\em arXiv preprint arXiv:2404.06395}.

\bibitem[Hu et~al., 2024b]{hu2024rdrec}
Hu, S., Tu, Y., Han, X., et~al. (2024b).
\newblock Rdrec: Rationale distillation for llm-based recommendation.
\newblock {\em arXiv preprint arXiv:2405.10587}.

\bibitem[Hu and Zhou, 2024]{hu2024llmevaluation}
Hu, T. and Zhou, X.-H. (2024).
\newblock Unveiling llm evaluation focused on metrics: Challenges and solutions.
\newblock {\em arXiv preprint arXiv:2404.09135}.

\bibitem[{\.I}rican et~al., 2024]{legal_bot}
{\.I}rican, B.~B., Sivri, M., Kokach, V., Koca{\c{c}}{\i}nar, B., and Akbulutl, F.~P. (2024).
\newblock Qbot: Domain-specific chatbots with retrieval-augmented generation and vector embedding for complex documentation queries.
\newblock In {\em 2024 Innovations in Intelligent Systems and Applications Conference (ASYU)}.

\bibitem[Jim{\'e}nez~Guti{\'e}rrez et~al., 2024]{memoryhipporag}
Jim{\'e}nez~Guti{\'e}rrez, B., Shu, Y., Gu, Y., Yasunaga, M., and Su, Y. (2024).
\newblock Hipporag: Neurobiologically inspired long-term memory for large language models.
\newblock {\em arXiv e-prints}.

\bibitem[Jin et~al., 2024]{jin2024llm}
Jin, H., Han, X., Yang, J., Jiang, Z., Liu, Z., Chang, C.-Y., Chen, H., and Hu, X. (2024).
\newblock Llm maybe longlm: Self-extend llm context window without tuning.
\newblock {\em arXiv preprint arXiv:2401.01325}.

\bibitem[Khattab et~al., 2023]{khattab2023dspy}
Khattab, O., Singhvi, A., Maheshwari, P., Zhang, Z., Santhanam, K., Vardhamanan, S., Haq, S., Sharma, A., Joshi, T.~T., Moazam, H., Miller, H., Zaharia, M., and Potts, C. (2023).
\newblock Dspy: Compiling declarative language model calls into self-improving pipelines.
\newblock {\em arXiv preprint arXiv:2310.03714}.

\bibitem[Lin et~al., 2024]{lin2024infinite}
Lin, B., Zhang, C., Peng, T., Zhao, H., Xiao, W., Sun, M., Liu, A., Zhang, Z., Li, L., Qiu, X., et~al. (2024).
\newblock Infinite-llm: Efficient llm service for long context with distattention and distributed kvcache.
\newblock {\em arXiv preprint arXiv:2401.02669}.

\bibitem[Liu et~al., 2024]{smantic_ex1}
Liu, C., Yuan, X., Yu, Z., and Wang, Y. (2024).
\newblock Texdc: Text-driven disease-aware 4d cardiac cine mri images generation.
\newblock In {\em Proceedings of the Asian Conference on Computer Vision (ACCV)}.

\bibitem[Mirzadeh et~al., 2024]{mirzadeh2024gsm}
Mirzadeh, I., Alizadeh, K., Shahrokhi, H., Tuzel, O., Bengio, S., and Farajtabar, M. (2024).
\newblock Gsm-symbolic: Understanding the limitations of mathematical reasoning in large language models.
\newblock {\em arXiv preprint arXiv:2410.05229}.

\bibitem[Su et~al., 2024]{rag_for_healthcare}
Su, C., Wen, J., Kang, J., Wang, Y., Pan, H., and Hossain, M.~S. (2024).
\newblock Hybrid rag-empowered multi-modal llm for secure healthcare data management: A diffusion-based contract theory approach.
\newblock {\em arXiv preprint arXiv:2407.00978}.

\bibitem[T~Wijesiriwardene, 2024]{smantic_ex4}
T~Wijesiriwardene, R.~W. (2024).
\newblock Exploring the abilities of large language models to solve proportional analogies via knowledge-enhanced prompting.
\newblock {\em arXiv preprint arXiv:2412.00869}.

\bibitem[van Schaik and Pugh, 2024]{van2024field}
van Schaik, T.~A. and Pugh, B. (2024).
\newblock A field guide to automatic evaluation of llm-generated summaries.
\newblock In {\em Proceedings of the 47th International ACM SIGIR Conference on Research and Development in Information Retrieval}, pages 2832--2836.

\bibitem[Wu et~al., 2024]{wu2024divide}
Wu, Z., Bai, H., Zhang, A., et~al. (2024).
\newblock Divide-or-conquer? which part should you distill your llm?
\newblock {\em arXiv preprint arXiv:2402.15000}.

\bibitem[Yang et~al., 2018]{yang2018hotpotqa}
Yang, Z., Qi, P., Zhang, S., Bengio, Y., Cohen, W.~W., Salakhutdinov, R., and Manning, C.~D. (2018).
\newblock Hotpotqa: A dataset for diverse, explainable multi-hop question answering.
\newblock {\em arXiv preprint arXiv:1809.09600}.

\bibitem[Yao et~al., 2024]{yao2024sarcasm}
Yao, B., Zhang, Y., Li, Q., and Qin, J. (2024).
\newblock Is sarcasm detection a step-by-step reasoning process in large language models?
\newblock {\em arXiv preprint arXiv:2407.12725}.

\bibitem[Zhang et~al., 2024a]{raft}
Zhang, T., Patil, S.~G., Jain, N., Shen, S., Zaharia, M., Stoica, I., and Gonzalez, J.~E. (2024a).
\newblock Raft: Adapting language model to domain specific rag.
\newblock {\em arXiv preprint arXiv:2403.10131}.

\bibitem[Zhang et~al., 2024b]{zhang2024selfcorrection}
Zhang, Y., Khalifa, M., Logeswaran, L., et~al. (2024b).
\newblock Small language models need strong verifiers to self-correct reasoning.
\newblock {\em arXiv preprint arXiv:2404.17140}.

\bibitem[Zhu et~al., 2024]{zhu2024llava}
Zhu, Y., Zhu, M., Liu, N., et~al. (2024).
\newblock Llava-phi: Efficient multi-modal assistant with small language model.
\newblock {\em arXiv preprint arXiv:2401.02330}.

\end{thebibliography}

\section*{\uppercase{APPENDIX}}
\label{appendix_label}
This section presents examples of inferences drawn from the experiments.
\subsection*{Distillation}
The input(Q,C,A):
\begin{small}
\begin{verbatim}
"""
Instruction: Given the question, 
the context 
and the expected answer bellow, 
provide relevant quotes from the 
context that support the answer.
your answer must be just the 
quotes, not the entire context.
format: 
##begin_quote##quote##end_quote##
for each quote.
do not add anything else other 
than the quotes.
Your turn:
Question: Unlike Xuzhou, where is Rugao 
under the adminstration of?
Context: Rugao () is a county-level city 
under the administration of Nantong, 
Jiangsu province, China, located in 
the Golden Triangle region on the 
northern (left) bank of the Yangtze River.
Xuzhou, known as Pengcheng in ancient 
times, is a major city in and the fourth
largest 
prefecture-level city of Jiangsu Province, 
China.  Its population was 8,577,225 
at the 2010 census whom 2,
623,066 lived in the built-up (or metro) 
area made of Quanshan, Gulou, Yunlong and 
Tongshan districts. 
It is known for its role as a transportation 
hub in northwestern Jiangsu, 
as it has expressways and railway links 
connecting directly to the provinces of 
Henan and Shandong, 
the neighboring port city of Lianyungang,
as well as the economic hub Shanghai.
Answer: Nantong
Quotes:    
""")
\end{verbatim}
\end{small}
And the generated quotes(R):
\begin{small}
\begin{verbatim}
"""
##begin_quote## Rugao () is a county-level
city under the administration of 
Nantong ##end_quote##
"""
\end{verbatim}
\end{small}

\subsection*{Quote Train Sample}
\textbf{The input(Q,C)}:
\begin{small}
\begin{verbatim}
"""
Instruction: Given the question
and the context
provide relevant quotes from the
context that support the answer.
your answer must be just the
quotes, not the entire context.
format:
##begin_quote##quote##end_quote##
for each quote.
do not add anything else other than
the quotes.
Question: What authority manages
the regional
passenger train service that
runs through 
the same junction as 
West Amesbury Branch Railroad?
Context: Rapido was the brand name
for the Canadian National Railway's 
(CN) express passenger train service 
in the Quebec City–Windsor Corridor. 
"Rapido" service was introduced on 
[..about 100 lines hide..] 
a shorter portion of the line.  
Before regional service was extended
to Denizli after track upgrades, 
three daily trains in each direction 
would operate as the Basmane-Nazilli
Regional.  Train service started in 2009 
after the completion of 
the Şirinyer Tunnel.
A rail replacement bus service uses 
buses to replace a passenger train 
service either on a 
temporary or permanent basis.  The 
train service that is replaced may be 
of any type such as light rail, tram, 
streetcar, commuter rail, regional rail 
or heavy rail, intercity 
passenger service.  
The rail service may be replaced if the 
line is closed due to rail maintenance, 
a breakdown of a train, a rail accident,
strike action, or if the rail service 
is not economically viable.
Quotes:
"""
\end{verbatim}
\end{small}
The \textbf{desired output(R)}:
\begin{small}
\begin{verbatim}
"""
##begin_quote## The Downeaster is a 145 mi
regional passenger train service, managed 
by the Northern New England Passenger Rail 
Authority (NNEPRA, created by the State of
Maine), and operated by Amtrak. ##end_quote##
##begin_quote## The West Amesbury Branch 
Railroad was a railroad that once led form 
Newton Junction, New Hampshire to Merrimac, 
Massachusetts.  Although the railroad does 
not exist, the Amtrak "Downeaster" line 
now passes through the Newton Junction 
station, which is now a pizza restaurant, 
and most of the railroad, is now a gravel 
walking trail. ##end_quote##
"""
\end{verbatim}
\end{small}

Table~\ref{tab:train_table} demonstrates an example of model performance before and after training in a test sample for the question: "Which film was produced first, Emil and the Detectives or Toy Story 2?"
\begin{table*}[]\footnotesize
\centering
\caption{Golden quote , before and after train quotes inferences from a test sample}
\label{tab:train_table}
\begin{tabular}{|c|c|c|}
\hline
\textbf{Gold}   & \textbf{Before} & \textbf{After}  \\ \hline
\begin{tabular}[c]{@{}l@{}}
\textbf{\#\#begin\_quote\#\#} Emil \\and the Detectives is a 1964 \\
film directed by Peter\\ Tewksbury based on the novel\\
by German author Erich\\ Kästner. \textbf{\#\#end\_quote\#\#}\\
\textbf{\#\#begin\_quote\#\#} \\Toy Story 2 is a 1999 American \\
computer-animated \\comedy film produced by Pixar \\
Animation Studios \\for Walt Disney Pictures.\\
\textbf{\#\#end\_quote\#\#}
\end{tabular}  
& 
\begin{tabular}[c]{@{}l@{}}
\textbf{\#\#begin\_quote\#\#}\\ 
"A Bug's Life", "Monsters, Inc.", \\
"Finding Nemo", "Cars",\\
"The Incredibles","Ratatouille"\\
\textbf{\#\#end\_quote\#\#}\\
\textbf{\#\#begin\_quote\#\#} "Toy Story 3" \\
(2010) is the third \\installment in Pixar's\\
"Toy Story" series, \\and the sequel to 1999's \\
"Toy Story 2".\\
\textbf{\#\#end\_quote\#\#}\\
\end{tabular}
& 
\begin{tabular}[c]{@{}l@{}}
\textbf{\#\#begin\_quote\#\#} Emil \\and the Detectives is a 1964 \\
film directed by Peter\\ Tewksbury based on the novel\\
by German author Erich\\ Kästner. \textbf{\#\#end\_quote\#\#}\\
\textbf{\#\#begin\_quote\#\#} \\Toy Story 2 is a 1999 American \\
computer-animated \\comedy film produced by Pixar \\
Animation Studios \\for Walt Disney Pictures.\\
\textbf{\#\#end\_quote\#\#}
\end{tabular}
\\ \hline

\end{tabular}
\end{table*}
\subsection*{Comparison: Quote x Context}

An example illustrating the performance comparison between using full context and extracted quotes.

\textbf{Question}:
\begin{small}
\begin{verbatim}
"""
Which Walt Disney Pictures
film was created first,
Finding Dory or The Wild Country?
"""
\end{verbatim}
\end{small}

\textbf{Context}: A `5086` characters context about Disney and Pixar films, attractions, and related productions, focusing on “Finding Nemo,” “Finding Dory,” and Disney’s broader entertainment ventures.

\textbf{Quotes}:
\begin{small}
\begin{verbatim}
"""
##begin_quote## The Wild Country is a 1970
American adventure film produced by Walt
Disney Pictures and directed by Robert Totten.
##end_quote##
##begin_quote## Finding Nemo is a 2003 
American computer-animated 
family film produced
by Pixar Animation Studios and released 
by Walt Disney Pictures. ##end_quote##
"""
\end{verbatim}
\end{small}

It’s fascinating to note that the Llama3b trained model delivers the same quotes as the golden standard in this case. A detailed comparison of the model’s performance, based on answers generated using either the provided context or the golden quotes, is presented in Table~\ref{tab:context_vs_quotes}.

\begin{table*}[]\footnotesize
\centering
\caption{Comparison of Q/A results : context x quotes}
\label{tab:context_vs_quotes}
\begin{tabular}{|c|c|c|}
\hline
\textbf{Model}   & \textbf{Contex Answer} & \textbf{Golden Quotes Answer}  \\ \hline
gpt3.5-turbo   
& \begin{tabular}[c]{@{}l@{}}Finding Nemo was created first.\end{tabular}
& \begin{tabular}[c]{@{}l@{}}The Wild Country\end{tabular}
\\ \hline
llama3.2:1b   
& \begin{tabular}[c]{@{}l@{}}Finding Dory is created first.\end{tabular}
& \begin{tabular}[c]{@{}l@{}}The Wild Country\end{tabular}
\\ \hline
llama3.2:3b   
& \begin{tabular}[c]{@{}l@{}}Finding Dory is created first.\end{tabular}
& \begin{tabular}[c]{@{}l@{}}The Wild Country\end{tabular}
\\ \hline
\end{tabular}
\end{table*}

\end{document}